\documentclass{esannV2}
\usepackage[dvips]{graphicx}
\usepackage[latin1]{inputenc}
\usepackage{amssymb,amsmath,array,amsfonts}
\usepackage{graphics} 
\usepackage{color}
\usepackage{booktabs}
\usepackage{subfigure}
\usepackage{cancel}
\usepackage{hyperref}

\usepackage[style=base]{caption}  
\DeclareCaptionFormat{side}{\hspace*{0.2cm}\parbox{5cm}{#1#2#3}}
\DeclareCaptionFormat{side2}{\hspace*{0cm}\parbox{12cm}{#1#2#3}}
\usepackage[rightcaption]{sidecap}
\sidecaptionvpos{figure}{c}

\definecolor{darkgreen}{rgb}{0,0.6,0.2}
\definecolor{mypink}{RGB}{219, 48, 122}
\definecolor{some_color}{RGB}{123, 20, 80}

\begin{document}
\title{Learning compressed representations of blood samples time series with missing data}

\author{Filippo Maria Bianchi, Karl \O{}yvind Mikalsen and Robert Jenssen
%
\thanks{This work was funded by the Norwegian Research Council FRIPRO grant no. 239844 \emph{Next Generation Learning Machines} and IKTPLUSS grant no. 270738 \emph{Deep Learning for Health}.}
%
\vspace{.3cm}\\
%
\textit{Machine Learning Group -- UiT the Arctic University of Norway}
}

\maketitle

\begin{abstract}
Clinical measurements collected over time are naturally represented as \emph{multivariate time series} (MTS), which often contain \emph{missing data}.
An \emph{autoencoder} can learn low dimensional vectorial representations of MTS that preserve important data characteristics, but cannot deal explicitly with missing data.
In this work, we propose a new framework that combines an autoencoder with the \emph{Time series Cluster Kernel} (TCK), a kernel that accounts for missingness patterns in MTS.
Via kernel alignment, we incorporate TCK in the autoencoder to improve the learned representations in presence of missing data. 
We consider a classification problem of MTS with missing values, representing blood samples of patients with surgical site infection. 
With our approach, rather than with a standard autoencoder, we learn representations in low dimensions that can be classified better.
\end{abstract}

\section{INTRODUCTION}

The application of machine learning and deep learning brought a significant impact in healthcare industry, improving diagnosis outcomes and changing the way of providing care to  patients~\cite{cheng2016risk}.
The main challenge that machine learning is asked to solve is to discover relevant structural patterns in clinical data, usually concealed and difficult to detect manually. 

An important fraction of electronic health records are clinical measurements collected from patients over time, which are represented as multivariate time series (MTS)~\cite{che2016recurrent}.
Several efforts have been devoted to learn informative and compact representations of MTS~\cite{che2017time}, not only to improve the quality of the analysis, but also to manage the large amounts of data necessary to train deep learning models~\cite{miotto2016deep}.
Furthermore, MTS are characterized by complex relationships across the variables and time that must be accounted in the analysis.
However, most methods are designed to treat vectorial data and they cannot be trivially extended to capture such relationships.

The autoencoder (AE) is a type of neural network originally conceived as a non-linear dimensionality reduction algorithm~\cite{Hinton504}, which has been further exploited to learn data representations in deep architectures~\cite{bengio2009learning}.
AEs have been adopted to map time series data into \textit{codes}, which are real-typed vectors lying in a lower dimensional space~\cite{langkvist2014review}.

Clinical measurements are often recorded at irregular frequencies that change for different patients, across variables, and over time. Hence, after discretizing time,  the resulting MTS end up containing missing values~\cite{mikalsen2016learning}.
Missing values follow patterns that reflect medical conditions of the patients or decisions of the doctors and, therefore, are important to be included in the analysis.
Since AE cannot process data containing missing values, those are usually replaced with imputation techniques that, however, cannot capture those patterns as they only fill blanks trying to introduce as less bias as possible. 
On the other hand, a recently proposed method, called Time series Cluster Kernel (TCK)~\cite{mikalsen2017time}, computes an unsupervised kernel similarity between MTS with missing data.
TCK leverages on the configurations of missingness patterns to improve the evaluation of the similarity. 

In this work, we propose a completely unsupervised approach for learning compressed representations of MTS in presence of missing data. 
Towards that end, we utilize the \textit{deep kernelized autoencoder} (dkAE)~\cite{kampffmeyer2017deep}, a recently proposed architecture that embeds the properties of a given prior kernel in the code representation of an AE through kernel alignment. 
By introducing TCK as prior kernel, we extend the dkAE framework to time series. 
Moreover, due TCK's properties, the relationships among the learned codes accounts for the presence of missing data, yielding a more discriminative representation of the data.

We apply our method to classify MTS of blood samples, relative to patients with site infections contracted after surgery and with a high percentage of missing data.
We compare the classification results obtained on the representations learned by a standard AE with the ones of a dkAE implementing the alignment to TCK.
Results indicate that the learned codes not only provide a compact vectorial representation, but the same classifier achieves better results when operates in our code space rather than in the input space.

\section{METHODS}


\subsection{Time series Cluster Kernel}

The \emph{Time series Cluster Kernel} \cite{mikalsen2017time} exploits the missing patterns in MTS to compute their similarities, rather than relying on imputation methods that may introduce strong biases.  
TCK implements an ensemble learning approach wherein the robustness to hyperparameters is ensured by joining the clustering results of many Gaussian mixture models (GMM) to form the final kernel. 
Hence, no critical hyperparameters must be tuned by the user.

To deal with missing data, the GMMs are extended using informative prior distributions \cite{Marlin:2012:UPD:2110363.2110408}.
The TCK matrix is built by fitting GMMs to the set of time series for a range of numbers of mixture components, to provide partitions with different resolutions that capture both local and global structures in the data.
To enhance diversity in the ensemble, each partition is evaluated on a random subset of attributes and segments, using random initializations and randomly chosen hyperparameters. 
This also provides robustness in the hyperparameters selection.
TCK is then built by summing (for each partition) the inner products between pairs of posterior distributions corresponding to different MTS.

\subsection{Autoencoder}

AEs simultaneously learn two functions.
The first one, \textit{encoder}, provides a mapping from an input domain, $\mathcal{X}$, to a code domain, $\mathcal{C}$, i.e., the hidden representation.
The second function, \textit{decoder}, maps from $\mathcal{C}$ back to $\mathcal{X}$.
In AEs with a single hidden layer, the encoding and decoding function are $\mathbf{c} = \phi(\mathbf{W}_E\mathbf{x} + \mathbf{b}_E)$ and $\mathbf{\tilde{x}} = \psi(\mathbf{W}_D\mathbf{h} + \mathbf{b}_D)$, where $\mathbf{x}$, $\mathbf{c}$, and $\mathbf{\tilde{x}}$ denote, respectively, a sample from the input space, its hidden representation (the \textit{code}), and its reconstruction. 
While $\phi(\cdot)$ is usually implemented as a sigmoid, in the case inputs are real-valued vectors, the squashing nonlinearity in $\psi(\cdot)$ can be replaced by a linear activation.
Finally, $\mathbf{W}_{E}$ and $\mathbf{W}_{D}$ are the weights and $\mathbf{b}_{E}$ and $\mathbf{b}_{D}$ the bias of the encoder and decoder, respectively.

To minimize the discrepancy between the input and its reconstruction, model parameters are learned by minimizing a reconstruction loss
\begin{equation}
    \label{eq:distortion}
    L_r(\mathbf{x}, \mathbf{\tilde{x}}) = \mathbb{E}\left\{\lVert \mathbf{x} - \mathbf{\tilde{x}} \rVert^{2} \right\} \; .
\end{equation}

By stacking more hidden layers an AE is capable of learning more complex representations by transforming inputs through multiple nonlinear transformations.
In its native formulation, an AE can process vectorial data and, therefore, a MTS is flattened into a uni-dimensional vector when fed to the AE. 
Since an AE process inputs of same lengths, missing are filled with numeric values.

\subsection{Deep Kernelized Autoencoder}

\begin{SCfigure}[1][t!]
\includegraphics[width=0.45\columnwidth, keepaspectratio,trim={0.5cm 0.1cm 0cm 0cm},clip]{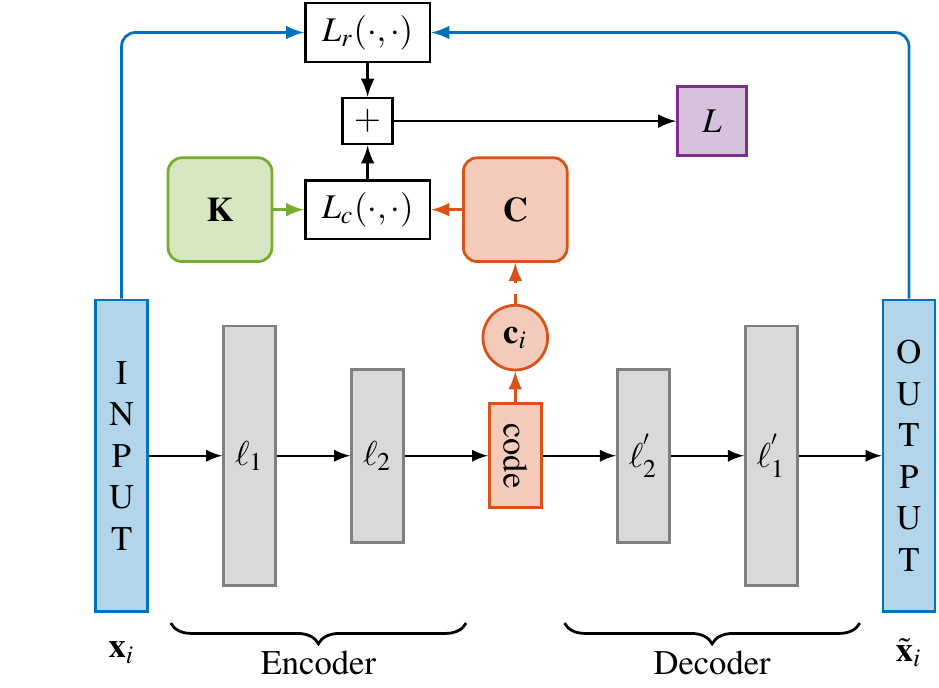}
\caption{Schematic illustration of dkAE architecture. The total loss function $L$ depends on two terms. 
First, $L_r(\cdot,\cdot)$, which computes the reconstruction error between true input $\mathbf{x}_i$ and output of dkAE, $\tilde{\mathbf{x}}_i$. The second term, $L_c(\cdot, \cdot)$, is the distance measure between the matrices $\mathbf{C}$ (computed as inner products of codes $\{ \mathbf{c}_i \}_{i=1}^{N}$) and the target prior kernel matrix $\mathbf{K}$.}
	\label{fig:kAE_arch}
\end{SCfigure}

A dkAE is trained by minimizing the loss function
\begin{equation}
    \label{eq:cost}
    L = (1-\lambda) L_r(\mathbf{x}, \mathbf{\tilde{x}}) + \lambda L_c(\mathbf{C}, \mathbf{K}),
\end{equation}
where $L_r(\cdot, \cdot)$ is the reconstruction loss in Eq.~\ref{eq:distortion} and $\lambda$ is a hyperparameter that balances the contribution of the two cost terms. 
If $\lambda=0$, $L$ becomes the traditional AE loss in Eq.~\ref{eq:distortion}.
$L_c(\cdot, \cdot)$ is the \textit{code loss} that enforces similarity between two matrices: $\mathbf{K} \in \mathbb{R}^{N \times N}$, the kernel matrix given as prior, and $\mathbf{C} \in \mathbb{R}^{N \times N}$, the inner product matrix of codes associated to input data.
A depiction of the training procedure is reported in Fig. \ref{fig:kAE_arch}.

$L_c(\cdot, \cdot)$ can be implemented as the normalized Frobenius distance between $\mathbf{C}$ and $\mathbf{K}$. Each matrix element $C_{ij}$ in $\mathbf{C}$ is given by 
$C_{ij}=\phi(\mathbf{x}_i) \cdot \phi(\mathbf{x}_j)$ and the code loss reads
\begin{equation}
\label{eq:regularization}
    L_c(\mathbf{C}, \mathbf{K}) = \Bigg{\lVert} \frac{\mathbf{C}}{\|\mathbf{C}\|_F} - \frac{\mathbf{K}}{\|\mathbf{K}\|_F} \Bigg{\rVert}_{F}.
\end{equation}

By minimizing the normalized Frobenius distance from TCK, we indirectly include in the codes the information it captures about the missingness patterns and we improve the quality of the learned codes in presence of missing data.

The dkAE model is trained using mini-batches.
Therefore, a training matrix $\mathbf{C}_m$ is generated from the codes associated to the elements in the $m$th mini-batch and distance $L_c$ is computed on the submatrix of $\mathbf{K}$ related to the entries in the mini-batch $m$.

\section{EXPERIMENTS}

We analyze blood measurements collected from patients undergoing a gastrointestinal surgery at University Hospital of North Norway in the years 2004--2012. 
Each patient in the dataset is represented by a MTS of blood samples extracted within $20$ days after surgery.
The MTS contain measurements of $10$ variables, which are alanine aminotransferase, albumin and alkaline phosphatase, creatinine, CRP, hemoglobine, leukocytes, potassium, sodium and thrombocytes.
We focus on a cohort of two classes of patients: the ones with and without surgical site infections.
Dataset labels are assigned according to International Classification of Diseases and NOMESCO Classification of Surgical Procedures, relative to patients with severe postoperative complications. 

Missing data in MTS correspond to measurements that are not collected for a given patient in one day of the observation period.
Patients with less than two measurements are excluded from the cohort. 
We ended up with $883$ MTS, of which $232$ are patients with infections.
The first $80\%$ of the datasets is used as training set and the rest as test set.

The dataset, the code implementing all the methods described in this paper, and a detailed description of experimental setup are publicly available\footnote{\url{https://github.com/FilippoMB/TCK_AE}}.

\subsection{Results}

To evaluate the effect of the alignment with TCK kernel, we compare the classification results obtained on the codes learned by standard AE and dkAE. 
Missing values are filled with three different imputation techniques: zero imputation (AE-z and dkAE-z), mean imputation (AE-m and dkAE-m) and last-value-carried-forward imputation (AE-l and dkAE-l).
The codes are classified by a $k$-NN with $k=3$ equipped with Euclidean distance.
We also consider the results yielded in the input space by a $k$NN with TCK similarity (TCK-i).

In Tab. \ref{tab:res} we report the mean and standard deviation of F1 score and area under the ROC curve (AUC) of the test set in 10 independent runs.
For AE and dkAE we also report the mean squared error (MSE) between the encoder input and the decoder output.
A low MSE of the reconstruction does not only guarantee to learn a better representation of the input, but it implies an accurate back-mapping from code to input space.
In both AE without kernel alignment and dkAE with zero imputation and last-value-carried-forward we obtain the best and worst classification performance, respectively.

For each imputation method, codes learned by dkAE are classified more accurately and the reconstruction error does not increase even if the codes are aligned with the prior kernel.
This demonstrate the importance of embedding into the codes the similarity information yielded by TCK, which captures missingness patterns.
Indeed, those patterns are ignored if one relies solely on imputation, whose purpose is to fill missing entries introducing as less bias as possible.
It is interesting to notice that the classification in the input space based on TCK similarity is slightly less accurate than the classification in the code space of dkAE.
Therefore, dkAE not only yields codes of reduced dimensionality that can be handled more easily and processed faster, but they are discriminated easier than the inputs themselves from a simple classifier.

\bgroup
\def\arraystretch{1.1} 
\setlength\tabcolsep{.2em} 
\begin{SCtable}[1][!ht]
\footnotesize
\centering
\caption{Reconstruction MSE and classification results of the codes learned by AE and dkAE. We also report the classification results in the input space using TCK as similarity. In AE and dkAE we apply three different imputations: zero imputation (z), mean imputation (m) and last value carried forward (l). Best results are highlighted in bold.}
\label{tab:res}
\begin{tabular}{l|ccc}
\cmidrule[1.5pt]{1-4}
\textbf{Method} & \textbf{MSE} & \textbf{F1} & \textbf{AUC} \\ 
\cmidrule[.5pt]{1-4}
\texttt{AE-z}   & 0.103$\pm$0.002           & 0.654$\pm$0.028           & 0.751$\pm$0.018 	        \\ 
\texttt{dkAE-z} & 0.096$\pm$0.001	        & \textbf{0.748}$\pm$0.017  & \textbf{0.813}$\pm$0.011  \\ 
\texttt{AE-m}   & 0.094$\pm$0.003 	        & 0.569$\pm$0.035	        & 0.7034$\pm$0.02  	        \\ 
\texttt{dkAE-m} & \textbf{0.091}$\pm$0.001  & 0.690$\pm$0.029       	& 0.773$\pm$0.018 	        \\ 
\texttt{AE-l}   & 0.136$\pm$0.002       	& 0.662$\pm$0.010     		& 0.764$\pm$0.006        	\\ 
\texttt{dkAE-l} & 0.128$\pm$0.000        	& 0.678$\pm$0.026     		& 0.763$\pm$0.016        	\\ 
\texttt{TCK-i}  & --    	                & 0.698$\pm$0.021	        & 0.776$\pm$0.012           \\ 
\cmidrule[1.5pt]{1-4}
\end{tabular}
\end{SCtable}
\egroup

\begin{figure*}[htp!]
\centering
	\subfigure
	{
	\includegraphics[width=3.5cm, height=2.8cm,trim={2.5cm 1cm 2cm 1.5cm},clip]{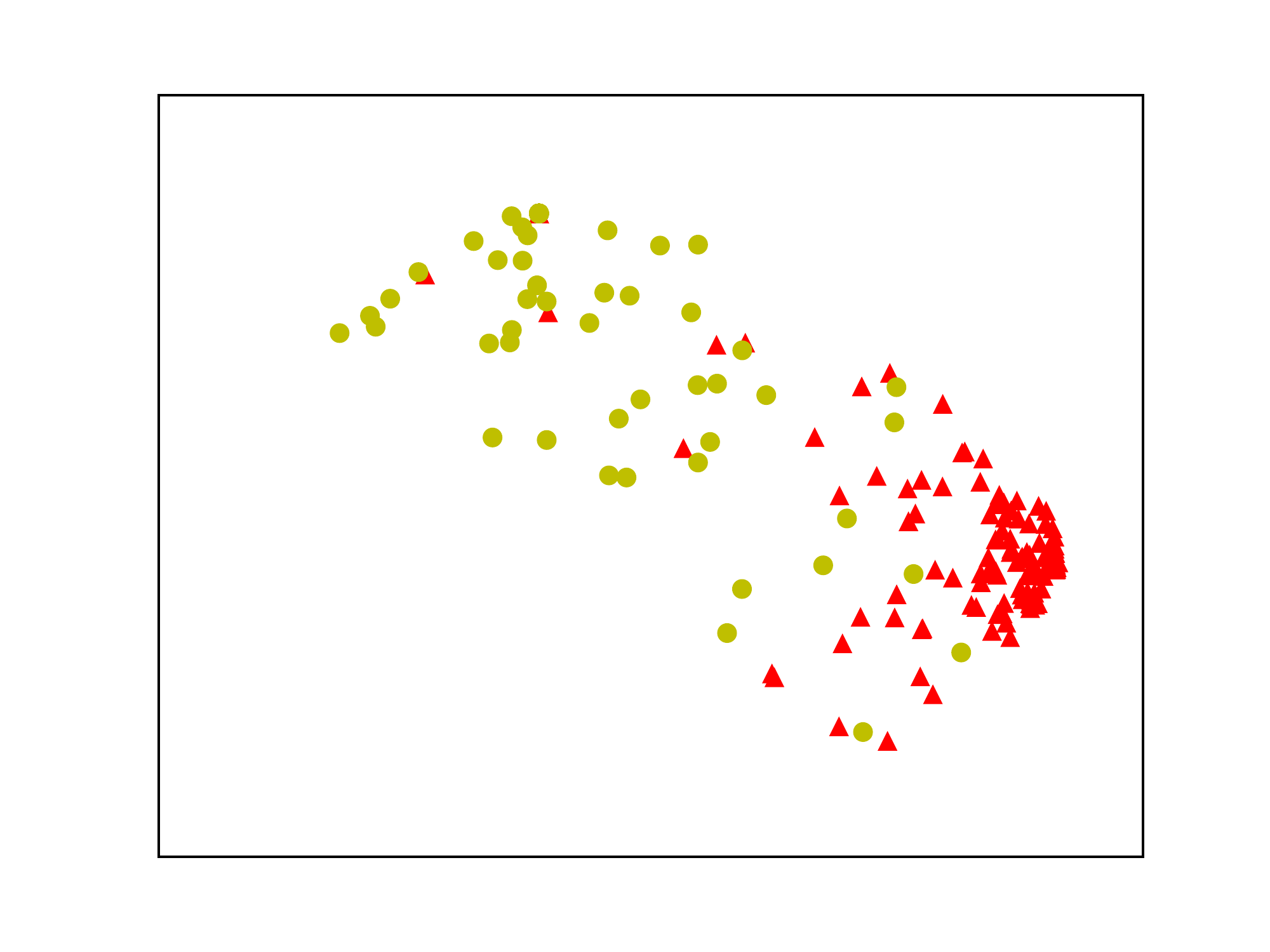}
	}
	~
	\subfigure
	{
	\includegraphics[width=3.5cm, height=2.8cm,trim={2cm 1cm 2cm 1.5cm},clip]{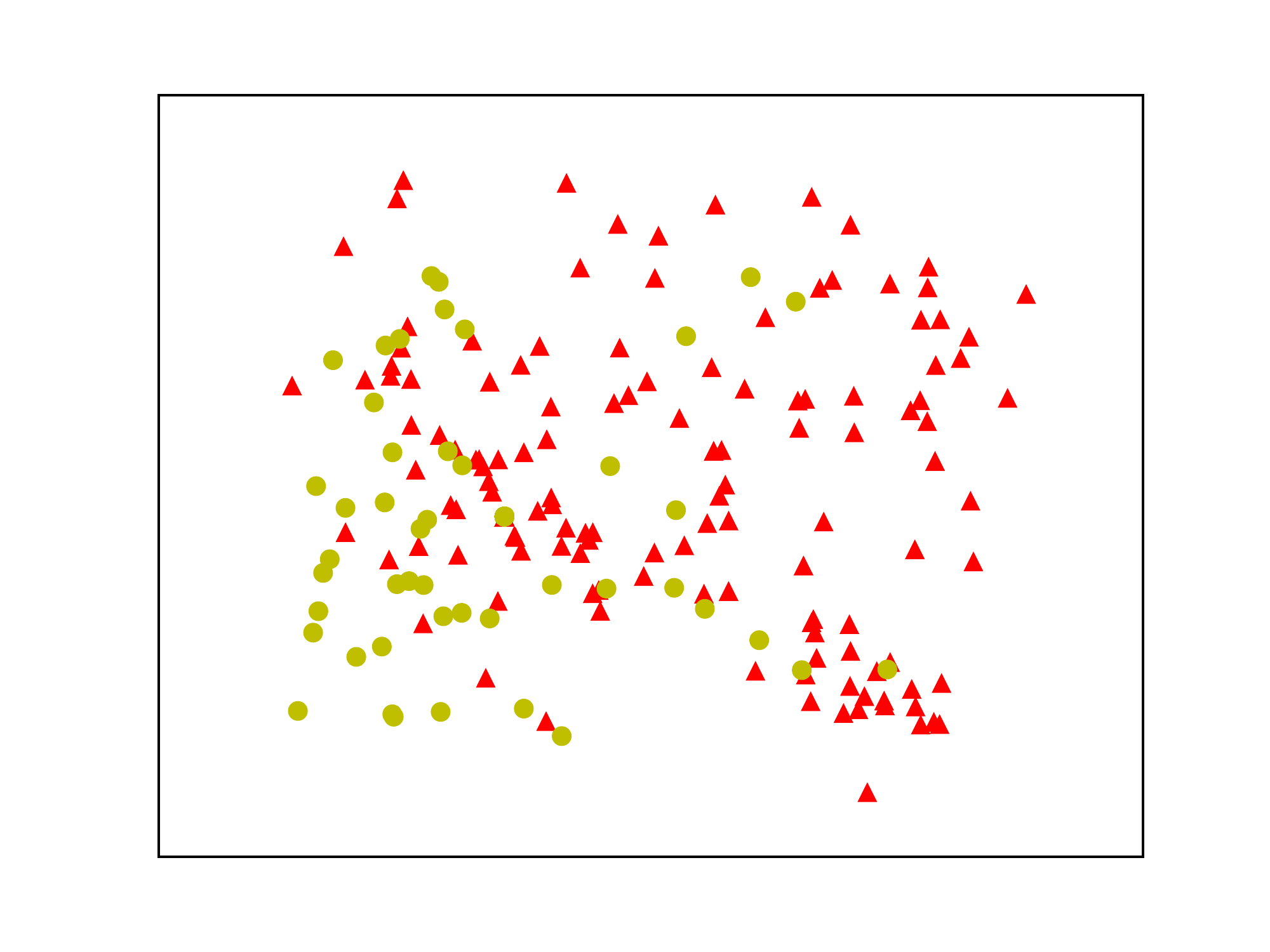}
	}
	~
	\subfigure
	{
	\includegraphics[width=3.5cm, height=2.8cm,trim={2.5cm 1cm 2cm 1.5cm},clip]{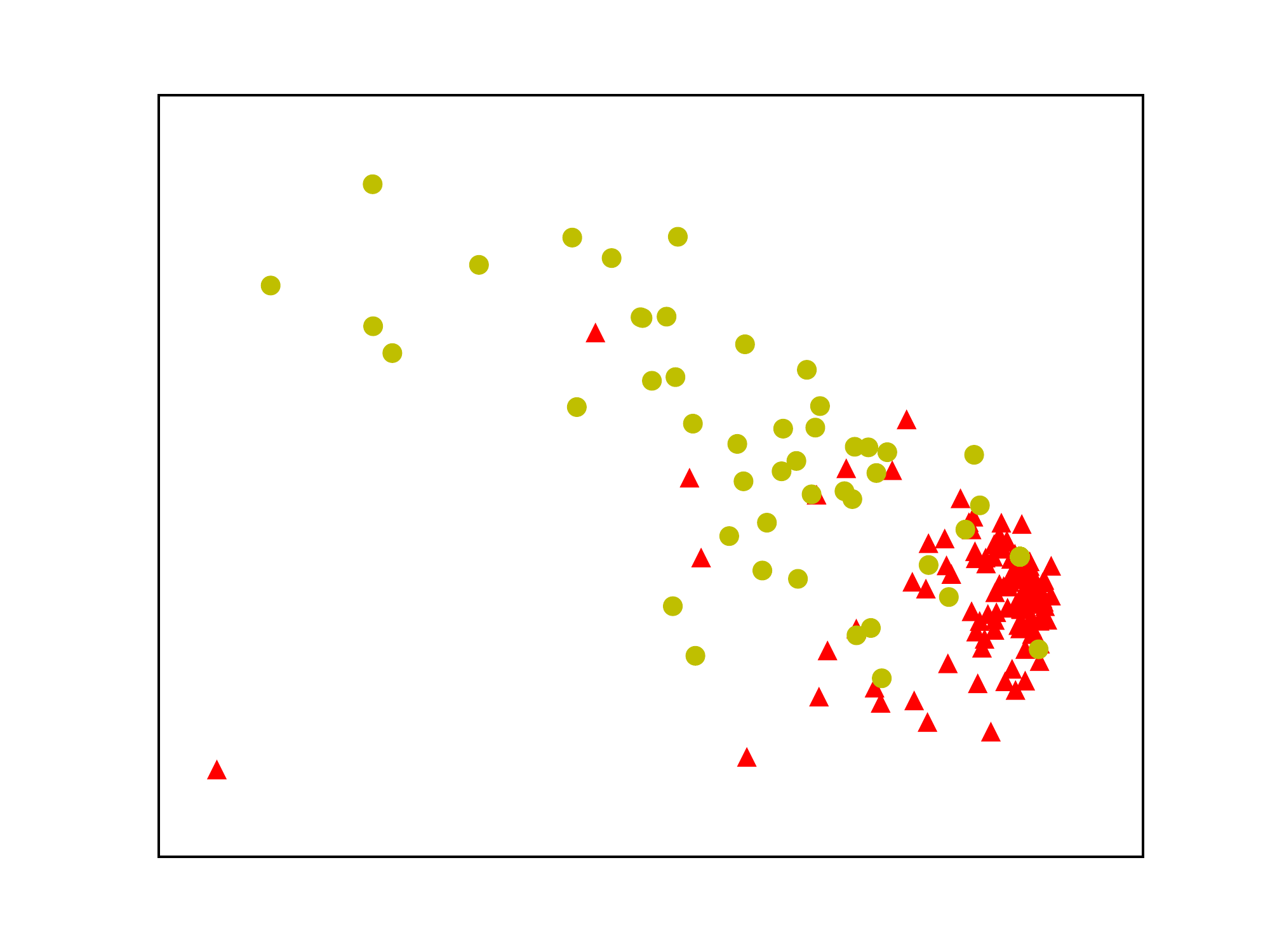}
	}
	\vspace{-4mm}
\caption{Projection of test set on the first two PCA components using (i) kPCA on the input space, (ii) PCA on AE code space, and (iii) PCA on dkAE code space. Yellow dots and red triangles represent infected and non-infected patients respectively.}
\label{fig:PCA}
\end{figure*}

In Fig. \ref{fig:PCA} we visualize the first two PCA components of the test set, both in input and in the code spaces.
We compute a linear PCA on the codes and on the TCK kernel matrix (this corresponds to compute kernel PCA in the input space using TCK as kernel).
Coloring depends on the ground truth label and we observe the two classes to be better separated in the code space of dkAE. 
Interestingly, in dkAE we notice the same structure yield by kPCA in the input space with TCK as kernel.
This demonstrate how the kernel alignment procedure successfully embed in the codes the properties of TCK, without compromising the precision of the decoder reconstruction.
We underline that by using an AE rather than kPCA we avoid performing a costly eigendecomposition and we also learn the inverse mapping from the code to the input space, provided by the decoder. 

\captionsetup{format=side2,font=footnotesize,labelfont=bf,labelsep=period}

\section{CONCLUSIONS}

In this paper, we proposed a novel approach for learning compressed vectorial representations of MTS with missing values, which are common in clinical records.
This is achieved by combining a deep kernelized Autoencoder with TCK, a similarity measure for MTS that accounts for missingness patterns.
We tackled the classification of blood samples from patients with postoperative infections, where data are MTS with a high percentage of missing data.
Our results showed that by aligning the codes in the AE to TCK kernel matrix, we embed into the representation important information relative to the missingess patterns in the data and improve the classification outcome.

\begin{footnotesize}


\bibliographystyle{unsrt}
\bibliography{biblio}

\end{footnotesize}


\end{document}